\pdfoutput=1

\documentclass[10pt,twocolumn,letterpaper]{article}
\usepackage{main}
\usepackage{times}
\usepackage{epsfig}
\usepackage{graphicx}
\usepackage{amsmath}
\usepackage{amssymb}
\usepackage{color}

\usepackage[breaklinks=true,bookmarks=false]{hyperref}

\cvprfinalcopy 

\def\cvprPaperID{3014}

\begin{document}

\title{Accurate Anchor Free Tracking}

\author{Shengyun Peng\\
College of Civil Engineering\\Tongji University\\
Shanghai, P.R. China\\
{\tt\small 1553983@tongji.edu.cn}
\and
Yunxuan Yu\\
Electrical and Computer\\Engineering Department, UCLA\\
Los Angeles, USA\\
{\tt\small yunxuan.yu@hotmail.com}
\and
Kun Wang\\
Electrical and Computer\\Engineering Department, UCLA\\
Los Angeles, USA\\
{\tt\small wangk@ucla.edu}
\and
Lei He\\
Electrical and Computer\\Engineering Department, UCLA\\
Los Angeles, USA\\
{\tt\small lhe@ee.ucla.edu}
}

\maketitle

\begin{abstract}
 Visual object tracking is an important application of computer vision. Recently, Siamese based trackers have achieved a good accuracy. However, most of Siamese based trackers are not efficient, as they exhaustively search potential object locations to define anchors and then classify each anchor (\textit{i.e.}, a bounding box). This paper develops the first Anchor Free Siamese Network (AFSN). Specifically, a target object is defined by a bounding box center, tracking offset and object size. All three are regressed by Siamese network with no additional classification or regional proposal, and performed once for each frame. We also tune the stride and receptive field for Siamese network, and further perform ablation experiments to quantitatively illustrate the effectiveness of our AFSN. We evaluate AFSN using five most commonly used benchmarks and compare to the best anchor-based trackers with source codes available for each benchmark. AFSN is $3\times-425\times$ faster than these best anchor based trackers. AFSN is also 5.97\% to 12.4\% more accurate in terms of all metrics for benchmark sets OTB2015, VOT2015, VOT2016, VOT2018 and TrackingNet, except that SiamRPN++ is 4\% better than AFSN in terms of Expected Average Overlap (EAO) on VOT2018 (but SiamRPN++ is $3.9\times$ slower). 
\end{abstract}

\begin{figure}[htbp!]
\begin{center}
\includegraphics[width=.46\textwidth]{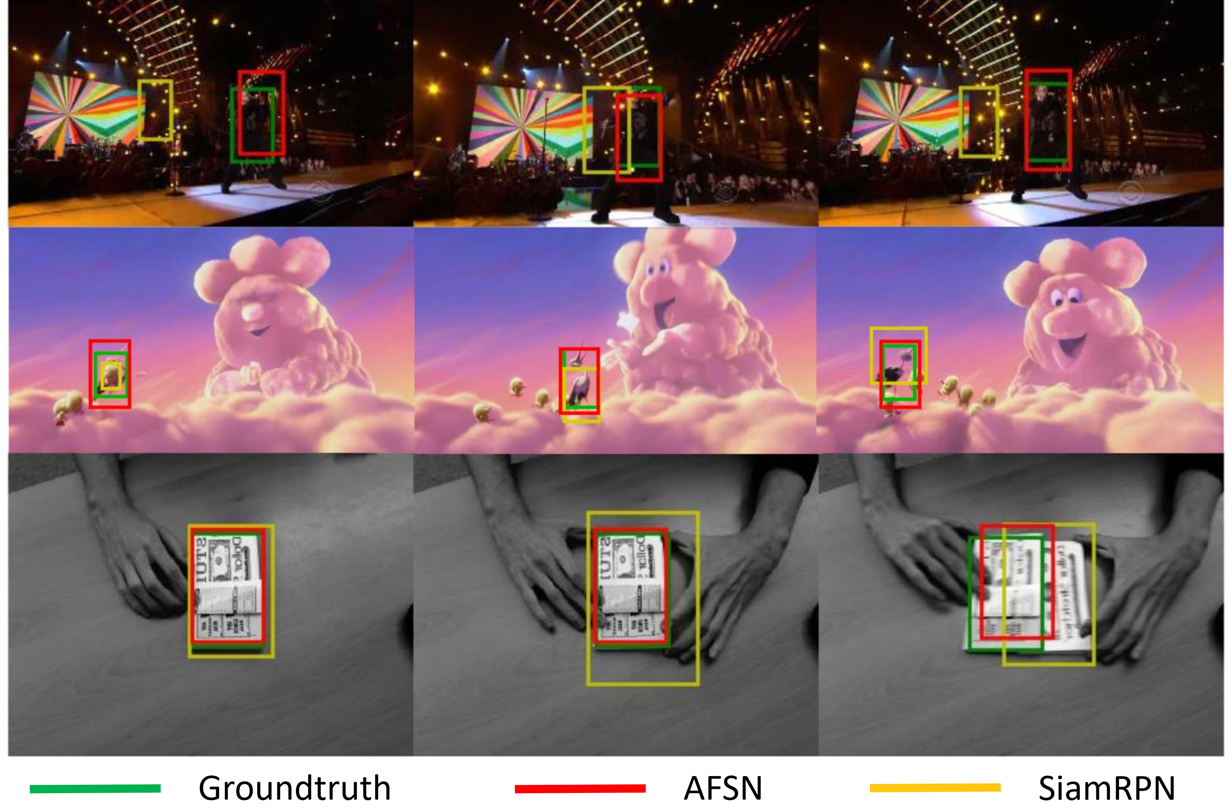}
\end{center}
  \caption{Comparisons of our tracker with SiamRPN. AFSN is able to resist the interference of similar objects, illumination variation, and predict a more precise bounding box than SiamRPN. When the ratio of length to width is abnormal, AFSN can still estimate the bounding box accurately.}
\label{fig:fig1}
\end{figure}
\section{Introduction}
Video object tracking, which locates an arbitrary object in a changing video sequence, powers many computer vision topics such as automatic driving and pose estimation. Liu \etal \cite{Liu2019} focus on the task of searching a specific vehicle that appears in the surveillance networks. Doellinger \etal \cite{Doellinger} use tracking methods to predict local statistics about the direction of human motion. A core problem in tracking is how to locate an object accurately and efficiently in challenging scenarios like background clutter, occlusion, scale variation, illumination change, deformation and other variations \cite{otb15}. 
\par Current trackers can be generally classified into two branches, \textit{i.e.}, generative and discriminative methods. Generative methods \cite{G1,G2,G3,G4,G5} consider tracking as a reconstruction problem and they maintain a template set online to represent the moving target. Discriminative trackers like MOSSE \cite{MOSSE}, Struck \cite{Struck}, CSK \cite{CSK}, KCF \cite{KCF} learn a classifier between foreground and background \cite{D1,D2,D3}. Correlation filter (CF) trackers can update online with current video due to its high efficiency. However, a clear deficiency of using data derived exclusively from current video results in learning a comparatively simple model. In contrast, trackers based on deep neural network aim to make full use of the entire tracking dataset \cite{MDNet}. Siamese networks, which track an object through similarity comparison, have developed into various versions in the tracking community \cite{SiamFC,DSiam,SiamRPN,SiamRPN++,DLtrack1,SINT,Attentional,SiamFT,DONOTLOSE,DSiamMFT}. 
\par Although these tracking approaches obtain balanced accuracy and speed, most of the successful Siamese trackers rely on the anchors generated before tracking. A tracking object is represented by an axis-aligned bounding box that encompasses the object. The localization of the object becomes an image classification of an extensive number of potential anchors. Since this method needs to enumerate all possible object locations and regress a normalized distance for each prospective bounding box, it is inefficient. Furthermore, it restricts the ability of proposing accurate bounding box when the ratio of length to width is abnormal.
\par To deal with this challenge, we propose the first tracker without using an anchor. First, our Anchor Free Siamese Network (AFSN) represents an object with merely a center point, a tracking offset and the object size. Compared with anchor-based trackers, it has a reduced complexity. Second, we model an object through the network inference result rather than modifying the object position and size with the bounding boxes proposed in advance. Only one estimation is conducted for each frame, with no need to classify each potential anchor. The inference efficiency is improved significantly. Third, our method leverages all images in the large scale supervised tracking datasets. Clearly, using videos from various categories can largely improve robustness. Ablation experiments also demonstrate the effectiveness of our AFSN.
\par To further improve the tracking quality, we test different network backbones. We find that the accuracy drops severely when the network backbone grows deeper. This problem has also been discovered in SiamDW \cite{SiamDW}. One reason is that these deeper and wider network architectures are mainly designed for image classification tasks, but not necessarily are optimal for tracking. We also reveal that a bigger network stride improves overlap area of receptive fields for two neighboring output score maps, but reduces tracking position precision, so the network stride needs to be optimized. In order to take full advantage of modern deep neural networks, we in this paper train 8 different backbones considering stride, receptive field, group convolution and kernel size. Section 4 gives a further analysis on the backbone design. The resulting AFSN outperforms a state-of-the-art tracker SiamRPN, as illustrated in Fig. \ref{fig:fig1}. 
\par We evaluate the proposed method using most commonly used datasets including OTB2015 \cite{otb15}, VOT2015 \cite{vot2015}, VOT2016 \cite{vot2016}, VOT2018 \cite{vot2018} and TrackingNet \cite{TrackingNet}. Our AFSN can achieve leading performance in all of the 5 datasets. Compared with SiamRPN, AFSN increases the precision rate and success rate by 0.93\% and 5.97\% on OTB2015, respectively. In terms of EAO (expected average overlap), it outperforms SiamRPN on VOT2015, VOT2016 and VOT2018 by 0.381, 0.372 and 0.398, respectively. Meanwhile, AFSN runs at 136 FPS (frames per second) on Titan Xp.
\par The contributions of this paper can be summarized in three-fold as follows: 1) We propose the Anchor Free Siamese Network (AFSN), which is the first anchor free end-to-end tracker trained with large scale dataset. 2) A quantitative analysis of the network architecture, especially receptive field and network stride, leads to the best network backbone for our AFSN. 3) Our tracker AFSN has balanced accuracy and robustness for five commonly used datasets. 
\par The rest of the paper is organized as follows. Section 2 introduces related work. Section 3 presents our AFSN, and Section 4 optimizes the network backbone. Section 5 performs experimental study, and Section 6 concludes the paper.

\begin{figure*}[htbp!]
\begin{center}
\includegraphics[width=.9\textwidth]{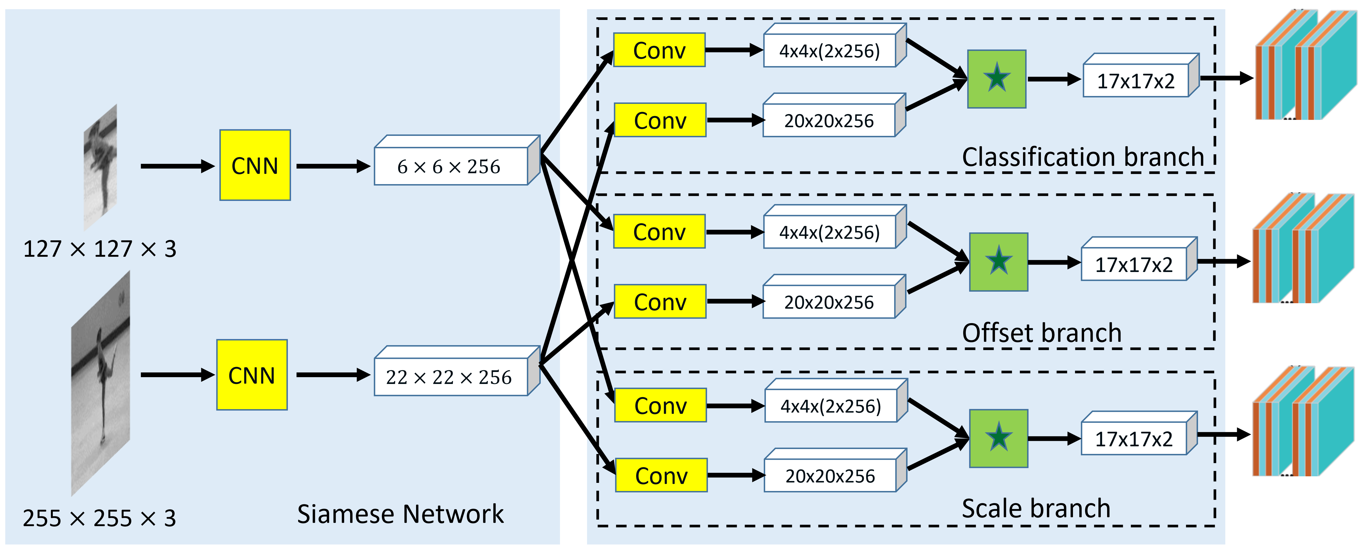}
\end{center}
   \caption{Main framework of Anchor Free Siamese Network (AFSN): On the left is the feature extraction subnetwork. Three branches, namely classification branch, offset branch and scale branch lies in the middle. These three branches are kused for classifying foreground and background, eliminating the deviation and predicting the object size. Then, depth-wise correlation is performed to obtain the final 2-layer score map, 2-layer tracking offset for $x$ and $y$ axis and 2-layer object size: width and height.}
\label{fig:network}
\end{figure*}

\section{Related work}
\subsection{Trackers based on Siamese network}
Recently, Siamese network has drawn widespread attention in tracking community because of its good accuracy and speed, and the capability to make full use of the tracking dataset during offline training. Basically, a Siamese network is used for comparing the exemplar and instance image pairs, and exporting the final results by a score map. SINT \cite{SINT} proposes learning a generic matching function for tracking, which can be applied to new tracking videos of previously unseen target objects. GOTURN \cite{DLtrack1} adapts the Siamese network to tracking and utilizes fully connected layers as fusion tensors. SiamFC \cite{SiamFC} introduces the correlation operator. Dense and efficient sliding window evaluation is achieved with a biliear layer that computes the cross correlation of its two inputs, namely instance branch and exemplar branch. SiamRPN \cite{SiamRPN} integrates a popular detection technique, namely region proposal network (RPN) with SiamFC. The tracker refines the proposal to avoid the expensive multi-scale tests. SiamRPN++ \cite{SiamRPN++} is a ResNet-driven Siamese tracker, which performs layer-wise and depth-wise aggregations. However, a tracking object is represented by an axis-aligned bounding box that emcompasses the object. The localization of the object becomes an image classification of an entensive number of potential anchors. Since this method needs to enumerate all possible object locations and regress a normalized distance for all propective bounding boxes, sliding window based object trackers are however a bit inefficient. 

\subsection{Dataset}
OTB2015 \cite{otb15} constructs a benchmark dataset with 100 fully annotated sequences to facilitate the performance evaluation. It is an extension of OTB2013 \cite{otb2013}, which contains 50 representative video sequences. The VOT \cite{vot2016,vot2017,vot2018} datasets are constructed by a novel video clustering approach based on visual properties. The dataset is fully annotated, all the sequences are labelled per-frame with visual attributes to facilitate in-depth analysis. The OTB \cite{otb2013,otb15}, ALOV \cite{ALOV} and VOT \cite{vot2016,vot2017,vot2018} datasets represent the initial attempts to unify the testing data and performance measurements of generic object tracking. Recently, GOT-10k \cite{GOT10k} has been proposed and is larger than most tracking datasets, which offers a much wider coverage of moving object. Several competitive trackers (MDNet \cite{MDNet}, SINT \cite{SINT}, GOTURN \cite{DLtrack1}) are trained on video sequences using OTB and VOT dataset. However, this practice has been prohibited in the VOT challenge. Thus, we train our network with the GOT-10k dataset, which differs from the video sequences in the benchmark. It is less likely for our model to over-fit the scenes and objects in the benchmark.

\subsection{Anchor free tracking}
Faster RCNN \cite{Fasterrcnn} generates region proposal within the detection network. It samples fixed shape anchors on a low resolution image and classifies each into foreground or background. SiamRPN \cite{SiamRPN} has adopted RPN into tracking scenario. The improved versions of SiamRPN, DaSiamRPN \cite{DaSiamRPN} and SiamRPN++ \cite{SiamRPN++} are all successful trackers. However, the enumeration of a nearly exhausted list of anchors is inefficient and requires extra post-processing. The tracking accuracy is also restricted by the pre-proposed fixed shape bounding boxes.
\par Keypoint estimation has some great applications in detection. CornerNet \cite{CornerNet} detects two bounding box corners as keypoints. ExtremeNet \cite{ExtremeNet} predicts the four corners and center points for all objects. CenterNet \cite{CenterNet} extracts a single center point per object without the need for post-processing. Since anchor free method solely generates the bounding box once in one inference time, its simplicity can boost the deep learning trackers. Different anchor free trackers represent the object using different techniques. Some represent the four corners, others represent the center point. This provides a variety of opportunities using anchor free in tracking. Anchor free has many successful applications in detection because of its efficiency and great performance. However, it has not been fully exploited in tracking. 

\section{Siamese tracking without anchors}
In this section, we describe the proposed AFSN framework in detail. As shown in Fig. \ref{fig:network}, the AFSN consists of a Siamese network for feature extraction. Three branches, namely classification branch, offset branch and scale branch are used for classifying foreground and background, eliminating the deviation and predicting the object size. Image pairs are fed into the proposed framework for end-to-end training.

\subsection{Bounding box center}
Our aim is to represent the object with the bounding box center. Scales and tracking offset are regressed directly from image features at the center location. Let $Y \in R^{W \times H \times 3}$ be an output score map of classification branch with width W and height H. Suppose $\hat Y_{\left(x_i, y_j, k\right)}$ is the value of point $\left(x_i, y_j\right)$ on the score map of the $k$th frame. A prediction $\hat Y_{\left(x_i, y_j, k\right)} = 1$ corresponds to the tracking object center, while $\hat Y_{\left(x_i, y_j, k\right)} = 0$ is the background. 
\par The classification labels $Y$ are designed to represent various foreground objects. Therefore, the groundtruth keypoints are designed to obey two-dimension normal distribution. The mean value is the center of bounding box. According to the three sigma rule \cite{ThreeSigma}, the probability for $X$ falling away from its mean by more than 3 standard deviation is at most 5\% if $X$ obeys the normal distribution. Thus, the standard deviation in our label is one sixth of the width and height:
\begin{equation}
    Y = \frac{1}{2\pi \sigma_1 \sigma_2} exp\left\{-\frac{1}{2} \left[ \frac{\left(x-\mu_1\right)^2}{\sigma_1^2}+\frac{\left(y-\mu_2\right)^2}{\sigma_2^2} \right]\right\}
\end{equation}
The response value intensifies with the increase of overlapping area between the exemplar and the instance. Hence, the score around the edge of bounding box should be lower than the center part. 
\par The training objective is a penalty-reduced pixel-wise logistic regression with focal loss \cite{focal}:
\begin{equation}
    L_{cls} = -\frac{1}{N} \sum
    \begin{cases}
    \left(1-\hat Y_{xyk}\right)^\alpha \log(\hat Y_{xyk}) & \text{$Y_{xyk}=1$}\\
    \begin{split}
        &\left(1-Y_{xyk}\right)^\beta \left(\hat Y_{xyk} \right)^\alpha\\  
        &\log(1-\hat Y_{xyk})
    \end{split}
    & \text{$Y_{xyk}=1$},
    \end{cases}
\end{equation}
where $\alpha$ and $\beta$ are hyper parameters of the focal loss, and $N$ is the number of frames in one epoch. We use $\alpha = 2$ and $\beta = 4$ in all our experiments, following Law and Deng \cite{CornerNet}.

\subsection{Tracking offset}
Since the input exemplar size, instance size and the output score map are $127\times{127}$, $255\times{255}$ and $17\times{17}$ separately, the stride of the network is 8. To eliminate the deviation and restore the gap, a tracking offset is added for each point on the score map. For the $i$ th point, tracking offsets $O_k=\left\{(\delta x_k^{(i)}, \delta y_k^{(i)})\right\}_{i=1}^n$ can be expressed as:
\begin{equation}
    O_k = \left( x_k/stride - \hat x_k, y_k/stride - \hat y_k \right)
\end{equation}
\par Then, the offset is trained with the L1 loss $L_{off}$. Each point on the score map is considered, which assists locating the bounding box even if the center peak deviates from the groundtruth.

\subsection{Scale estimation}
To estimate the size of an object is equivalent to regress the object size $S_k=(x_{k2}-x_{k1}, y_{k2}-y_{k1})$ in each frame. To make sure the estimation falls in the positive region, we represent the size with $\alpha_{k}$ and $\beta_{k}$ as:
\begin{equation}
    S_k = \left(e^{\alpha_{k}},e^{\beta_{k}}\right)
\end{equation}
\par Then, a simple prediction is conducted. The L1 loss for the scale estimation at the bounding box center is:
\begin{equation}
    L_{scl} = \frac{1}{N}\sum_k\left[\left|\hat \alpha_{P_k} - \alpha_k \right| + \left|\hat \beta_{P_k} - \beta_k \right|\right]
\end{equation}

\begin{figure}[htbp!]
\begin{center}
\includegraphics[width=.45\textwidth]{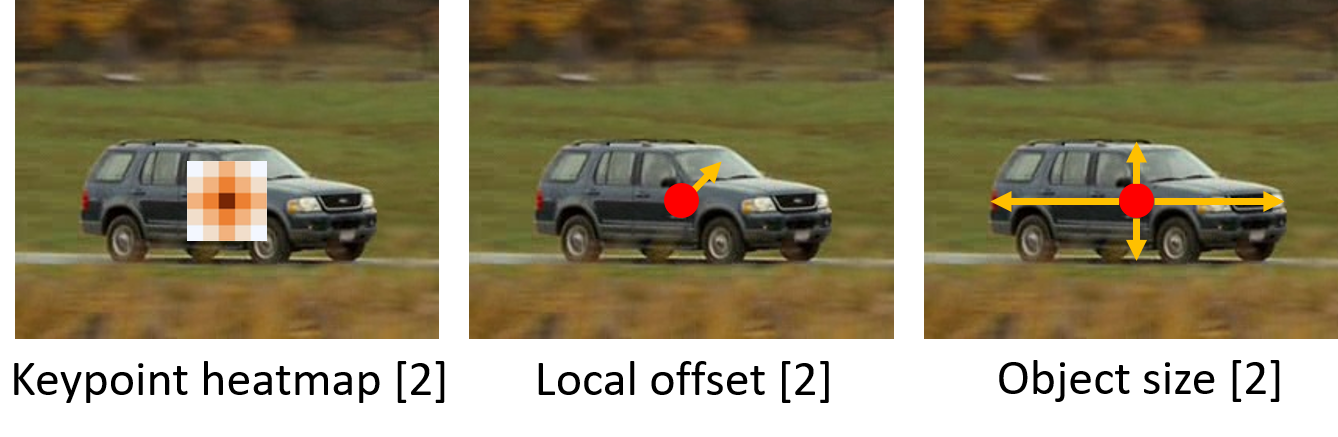}
\end{center}
  \caption{The outputs of the proposed anchor free Siamese network: score map, tracking offset and object size.}
\label{fig:output}
\end{figure}

\begin{figure*}[htbp!]
\begin{center}
\includegraphics[width=1.0\textwidth]{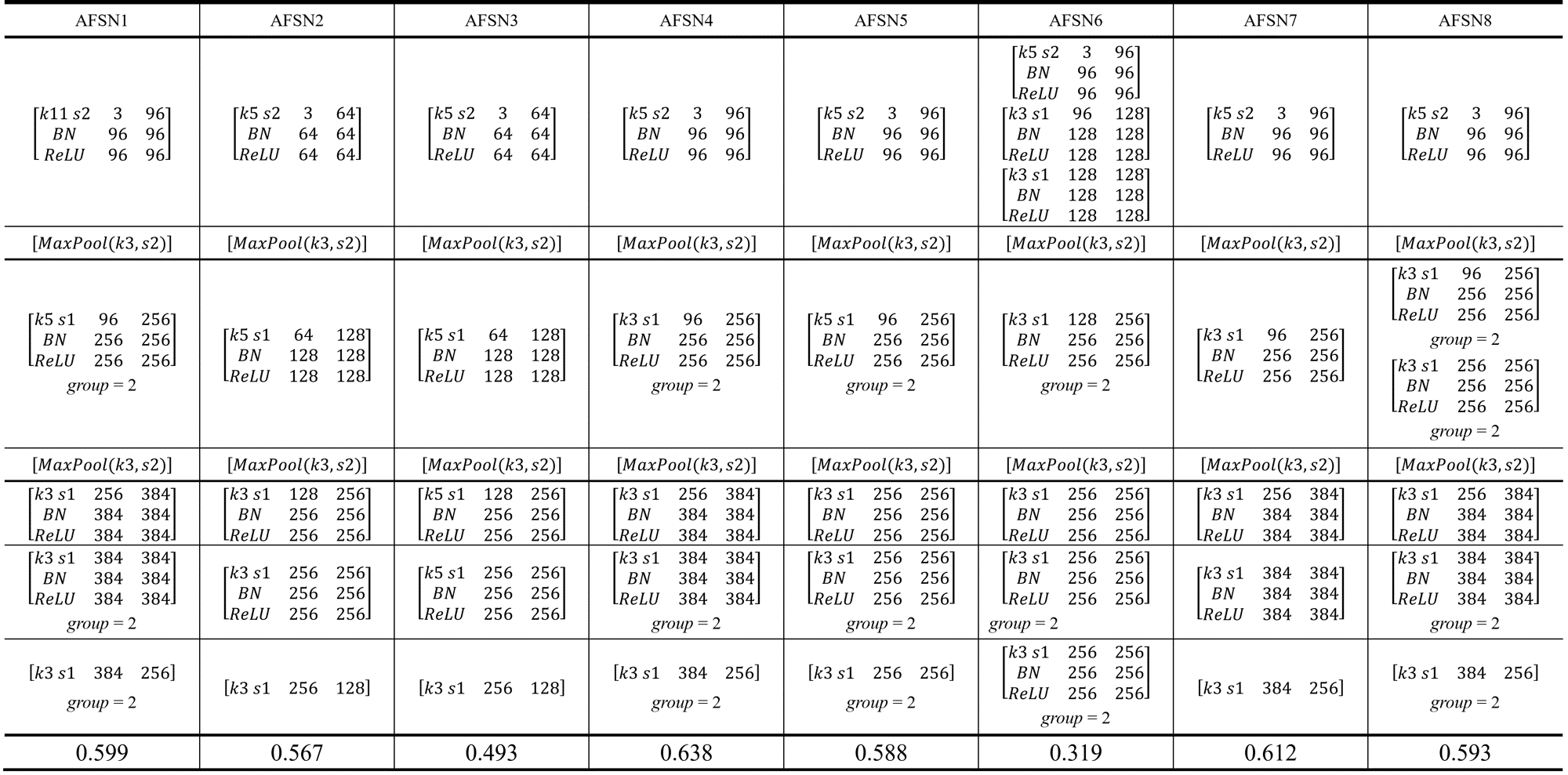}
\end{center}
  \caption{Architectures of designed backbone networks for AFSN. In this architecture list, $k$ is the kernel size, $s$ is the general convolution stride and $group$ is the number of group convolution. on the bottom is the success rate tested on OTB2015 benchmark.}
\label{fig:backbone}
\end{figure*}

\subsection{Training}
The feature extraction subnetwork is fully-convolutional. The search of optimum network architecture is presented in Section 4. Two branches compose the subnetwork. The template branch receives the exemplar patch (denoted as $z$). The search branch receives the full-scale instance patch (denoted as $x$). The two feature extraction branches share the same parameters. Thus, the same types of features can be compared in the following network. Let $L_t$ represent the extraction operator $\left(L_tx\right)[u]= x[u-t]$, and the definition of fully convolution within stride $k$ can be defined as:
\begin{equation}
    h\left(L_{kt}x\right) = L_th\left(x\right)
\end{equation}
\par Correlation operator is a batch processing function, which compares the Euclidean distance or similarity metric between $\phi(z)$ and $\phi(x)$. Note that $\phi(z)$ and $\phi(x)$ denote the outputs of template and search branches, respectively. Combining deep features in a higher dimension is equivalent to dense sampling around the bounding box and evaluating similarity after each feature extraction. However, the former method is more efficient due to the smaller scale of higher dimension feature. For convenience, let $u\left(\phi(z), \phi(x)\right)$ denote the output of correlation function. 
\par Since no normalization for offset and scale is included, the overall loss function is:
\begin{equation}
loss = L_{cls} + \lambda_{off} L_{off} + \lambda_{scl} L_{scl},
\label{loss}
\end{equation}
where $\lambda_{off}$ and $\lambda_{scl}$ are two hyper-parameters to balance the three parts. In our experiment, we set $\lambda_{off} = 0.1$ and $\lambda_{scl} = 4$. Only this single network is used to predict the bounding box center $\hat{Y}_k$, tracking offset $\hat{O}_k$ and object size $\hat{S}_k$

\subsection{Tracking}
At inference time, the points with the highest response score are extracted in the score map. In order to avoid noises or sudden changes in the background, we also apply a hanning window on the final score map. Suppose $\hat P_k = (\hat x_k, \hat y_k)$ is the predicted center point in the $k$ th frame. Combining the regressed tracking offset $\hat O_k = (\delta \hat x_k, \delta \hat y_k)$ and the object size $\hat S_k = (\hat w_k, \hat h_k)$, the estimated bounding box can be expressed as:
\begin{equation}
    \left(\hat x_k + \delta \hat x_k - \hat w_k / 2, \hat y_k + \delta \hat y_k - \hat h_k / 2, \hat w_k, \hat h_k\right) 
\end{equation}
A more explicit way of illustrating the Siamese output is shown in Figure \ref{fig:output}.

\section{Network backbone}
This section presents the process of optimizing the network backbone for the proposed AFSN tracker. Stride, receptive field, group convolution and kernel size are the impact factors of different networks. For a faster network searching process, the backbone networks are trained by 40\% GOT-10k dataset \cite{GOT10k} with 20 epochs. 
\par With the size of the input image and output score map, the stride of Siamese trackers can be calculated as:
\begin{equation}
    \frac{instance - exemplar}{score map - 1} = stride
    \label{eq:stride}
\end{equation}
The aggregation of different kernel size controls the region of the receptive field. A larger receptive field provides greater image context, but shallow features like color, shape will be lost. A smaller receptive field focuses on several particular parts on objects, but it cannot capture the structure of target objects. From Eq. \ref{eq:stride}, we can find that if we increase the receptive field, the score map size will decrease because more information are contained in one convolution. Then, the stride will increase, leading to a larger gap between two exemplar images. The final tracking results are generally correct around the target object, but the accurate localization and scale estimation cannot be achieved. If the receptive field decreases in order to capture the detail features, the gap will also decrease. However, once the bounding box deviates from the tracking object, it shows less robustness to relocate the object. Although the predicted scale will be more accurate in this scenario, the accuracy will not increase due to the poor robustness, In Siamese tracking, template image is not updated online, which further decreases the accuracy. This is a common contradictory in Siamese network, and it also explains why the backbone networks in Siamese trackers are relatively shallow. 
\par To optimize and search for the best backbone network, 8 different backbones are trained as shown in Fig. \ref{fig:backbone}. Network stride affects the overlap area of receptive fields for two neighboring output score maps. The proposed AFSN prefers a relatively small network stride, which is around 7 to 9 (AFSN1 \textit{vs.}AFSN6 \textit{vs.}AFSN8). In these cases, the depth of the shallow layers largely affect the success rate. Since shallow features like color and shape can be applied on several similar background objects, there is no need to extract more shallow features. Therefore, the receptive field is better set at 70\% to 80\% of the exemplar image. Group convolution separates different channels to different kernels, which increases the robustness of the tracking (ASFN4 \textit{vs.}AFSN7). It also decreases the computation amount. More channel numbers extract more features and offer more similarity information to compare, the optimum channel numbers for deeper layers is 256 (AFSN1 \textit{vs.}AFSN2 \textit{vs.}AFSN3).

\section{Experiments}
This section presents the results of our anchor free Siamese network on five challenging tracking datasets, \textit{i.e.}, OTB2015, VOT2015, VOT2016, VOT2018 and TrackingNet. All the tracking results are compared with the state-of-the-art trackers using the reported results to ensure a fair comparison.
\subsection{Implementation details}
The parameters in the proposed AFSN are trained from scratch, and the overall training objective optimizes the loss function in Eq. \ref{loss} with SGD. There are totally 50 epochs conducted and the learning rate decreases in the log space from $10^{-3}$ to $10^{-6}$. Since the loss in tracking offset occupies most of the loss in the first phase of training, we set a cut off at $10^{-8}$ for the offset loss. We extract image pairs from GOT-10k dataset for training and test on OTB and VOT dataset to verify the feasibility and efficiency of our model. The template image is cropped centering on the foreground object with size of $A \times A$:
\begin{equation}
    \left(w+p\right) \times \left(h+p\right) = A^2,
\end{equation}
where $w,h$ are the target bounding box width and height and $p=\frac{w+h}{2}$. For the template branch, $A$ is 127, and for the instance branch, $A$ is 255. 

\subsection{Ablation experiments}
\begin{figure}[htbp!]
\begin{center}
\includegraphics[width=.46\textwidth]{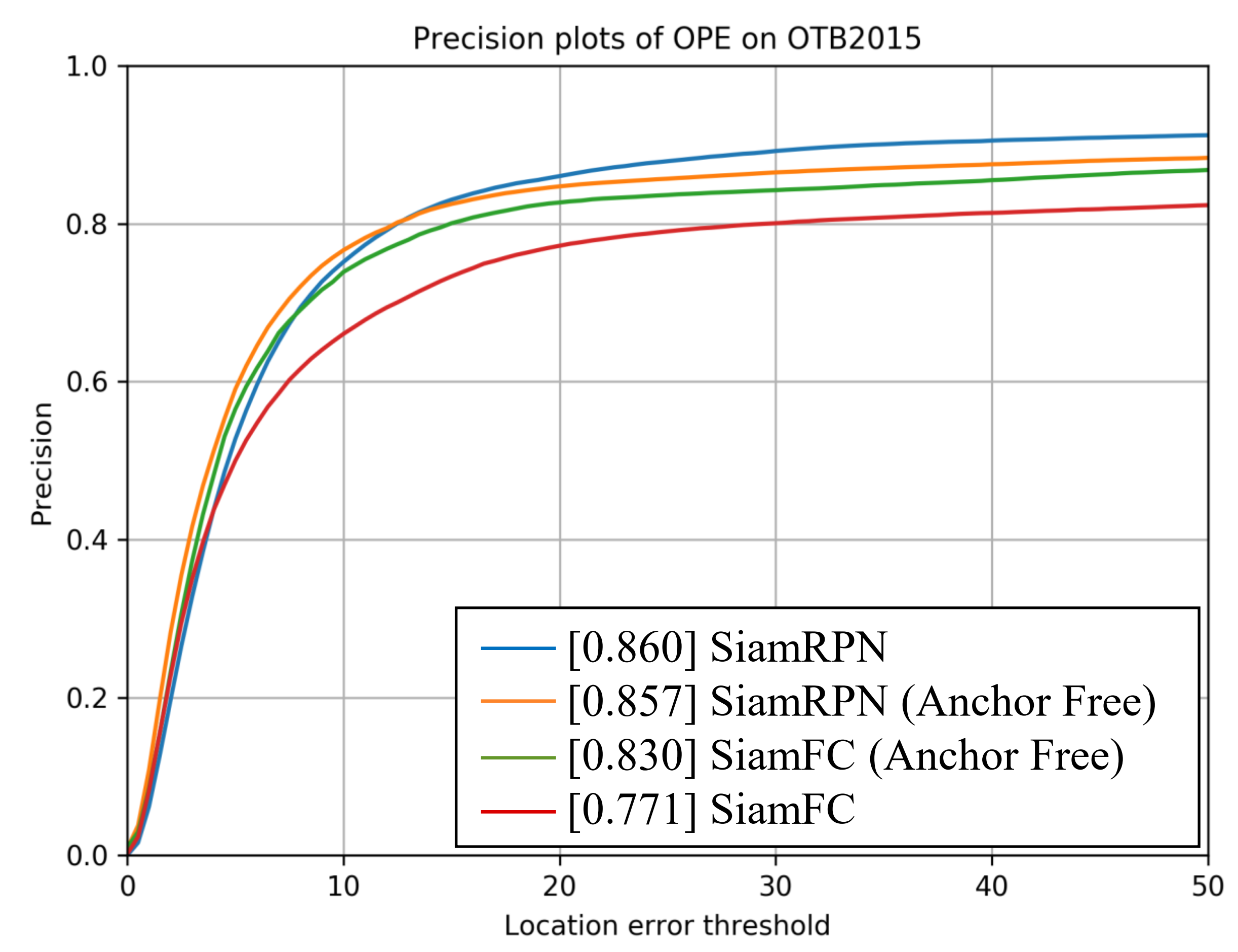}
\end{center}
  \caption{Ablation experiments: precision plot on OTB2015}
\label{fig:ablation1}
\end{figure}

\begin{figure}[htbp!]
\begin{center}
\includegraphics[width=.46\textwidth]{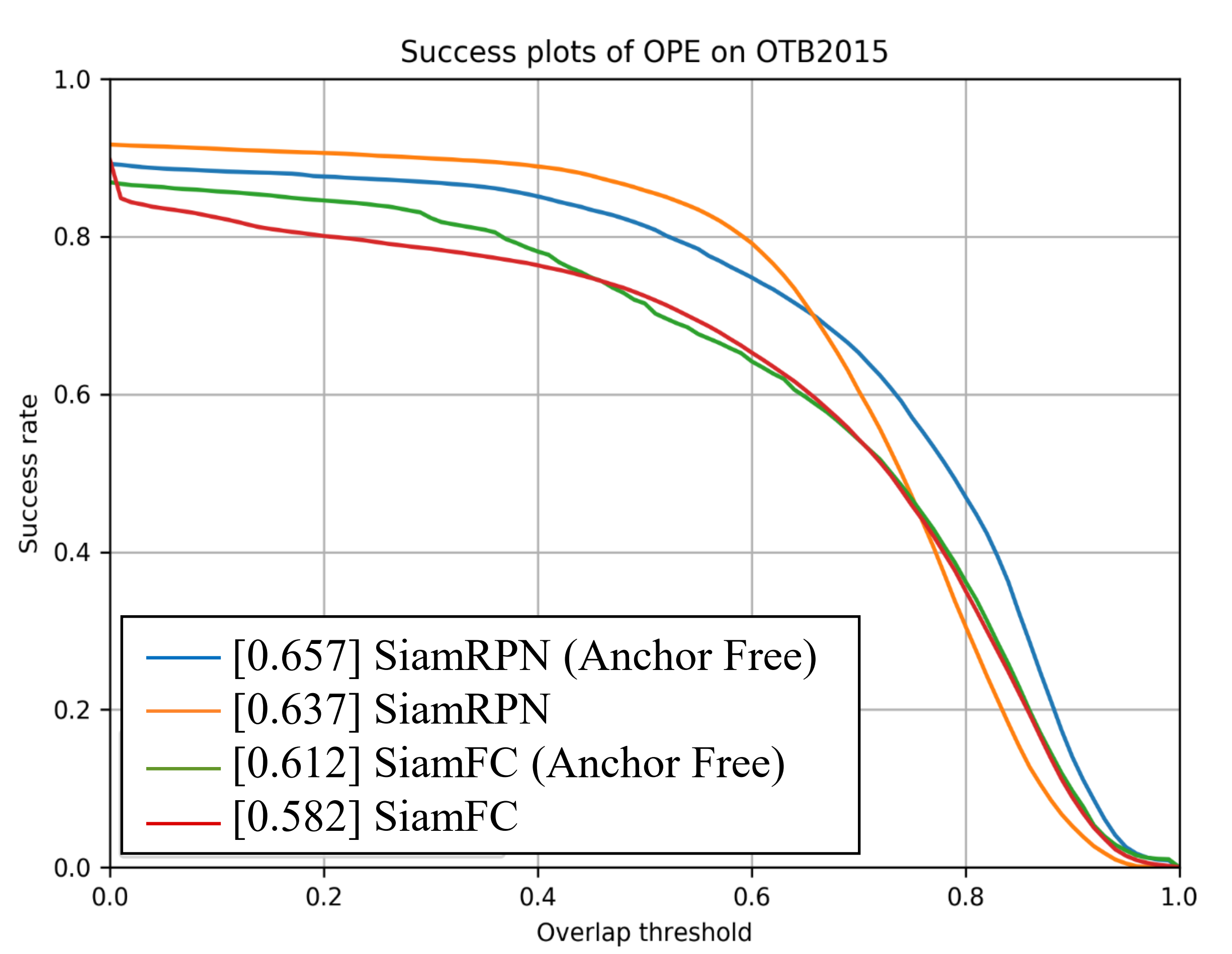}
\end{center}
  \caption{Ablation experiments: success plot on OTB2015}
\label{fig:ablation2}
\end{figure}

To investigate the impact of anchor free method, we change the training label for SiamFC and SiamRPN without changing the input and hyper-parameters of the original models. SiamFC finds the best tracking scale by enumerating three potential ratio: $1.025^{-1,0,1}$. We represent it with a new score map following the two-dimension normal distribution. The size can be directly predicted according to the response. Through combining the original model with the newly designed label, the precision rate and the success rate on OTB2015 grow 7.65\% and 5.15\%, as shown in Fig. \ref{fig:ablation1} and \ref{fig:ablation2}. We also apply this approach to SiamRPN. The precision rate drop a little, mainly because the tracking offset is not included. The position estimation may drift caused by the locations of the pre-proposed anchors. Even though no offset is incorporated, the success rate increases 3.14\%. The results demonstrate that the anchor free design can improve the tracking performance.

\subsection{Results on OTB2015}
\textbf{OTB-2015 Benchmark} The standardized OTB benchmark provides a fair test for both accuracy and robustness. The benchmark \cite{otb15} considers the precision plot and the success plot of one path evaluation (OPE). The precision plot considers the percentage of frames in which the estimated locations are within a given threshold distance of the target bounding box. The definition of average success rate is that a tracker is successful in a given frame if the intersection-over-union between its estimation and the groundtruth is above a certain threshold. 
\par
\begin{figure}[htbp!]
\begin{center}
\includegraphics[width=.46\textwidth]{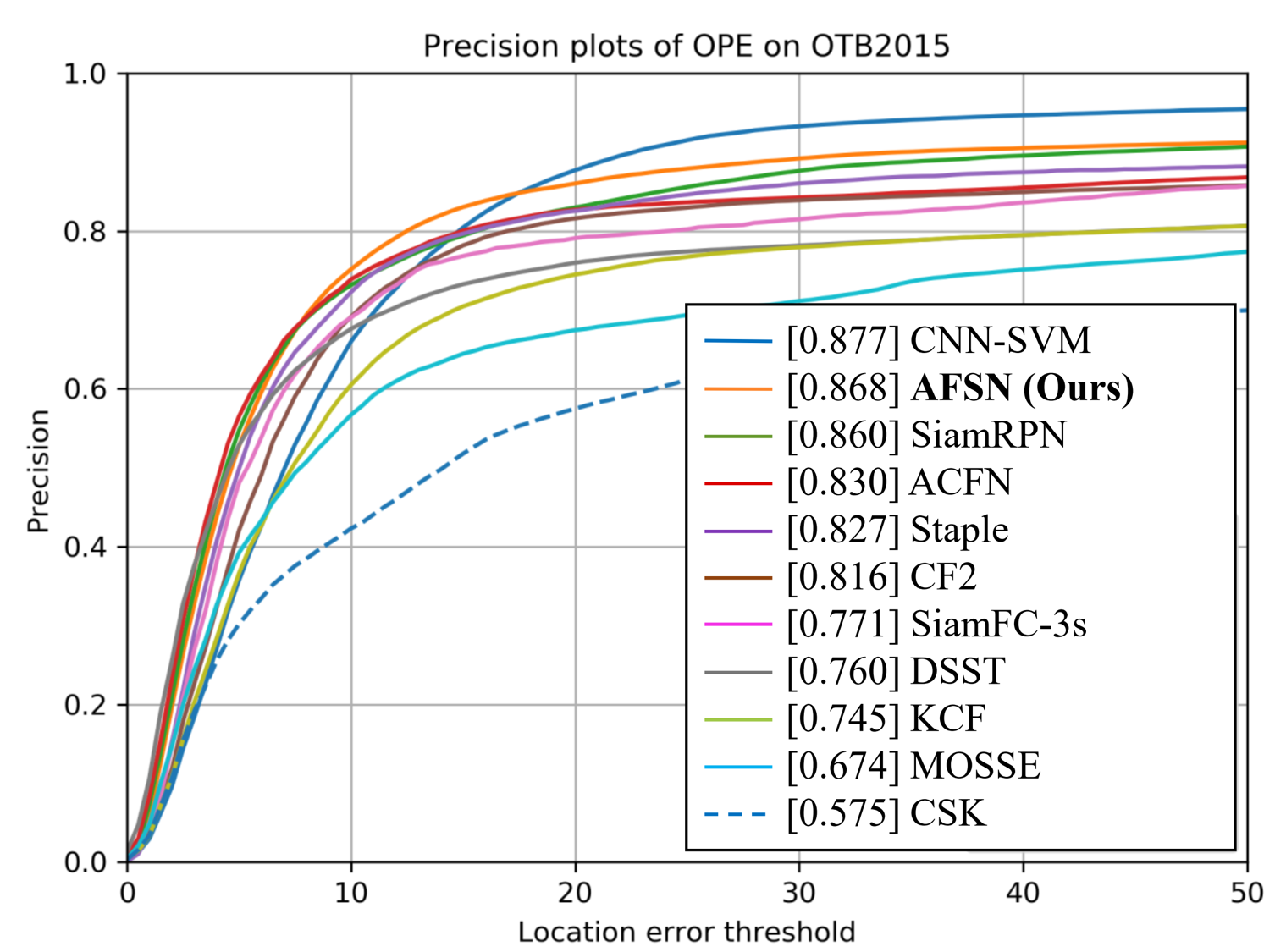}
\end{center}
  \caption{Precision plot of OTB2015}
\label{fig:precision}
\end{figure}

\begin{figure}[htbp!]
\begin{center}
\includegraphics[width=.46\textwidth]{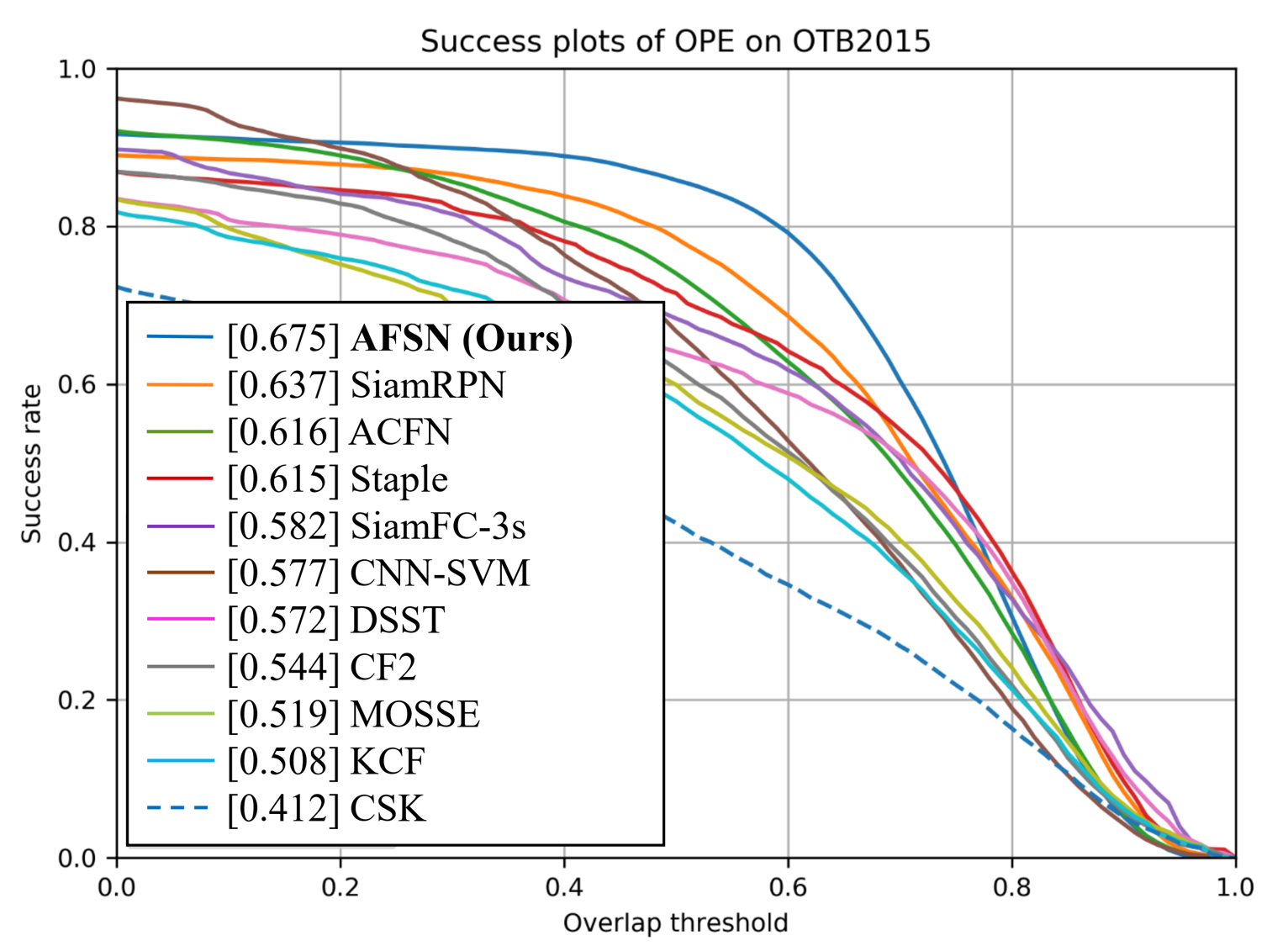}
\end{center}
  \caption{Success plot of OTB2015}
\label{fig:success}
\end{figure}

 We compare our anchor free Siamese tracker on the OTB2015 with the state-of-the-art trackers including SiamRPN \cite{SiamRPN}, ACFN \cite{ACFN}, Staple \cite{staple},SiamFC \cite{SiamFC}, CNN-SVM \cite{CNNSVM}, DSST \cite{dsst}, CF2 \cite{cf2}, MOSSE \cite{MOSSE}, KCF \cite{KCF}, CSK \cite{CSK}. Fig. \ref{fig:precision} and \ref{fig:success} show that our tracker produces leading results. Compared with the recent SiamRPN \cite{SiamRPN}, the precision rate and success rate increase 0.93\% and 5.97\%, respectively. 

\subsection{Results on VOT2015}
The VOT2015 dataset consists on 60 sequences \cite{vot2015}. The overall performance is evaluated using Expected Average Overlap (EAO), which takes account of both accuracy and robustness. Besides, the speed is evaluated with a normalized speed Equivalent Filter Operations (EFO).
\par
\begin{figure}[htbp!]
\begin{center}
\includegraphics[width=.46\textwidth]{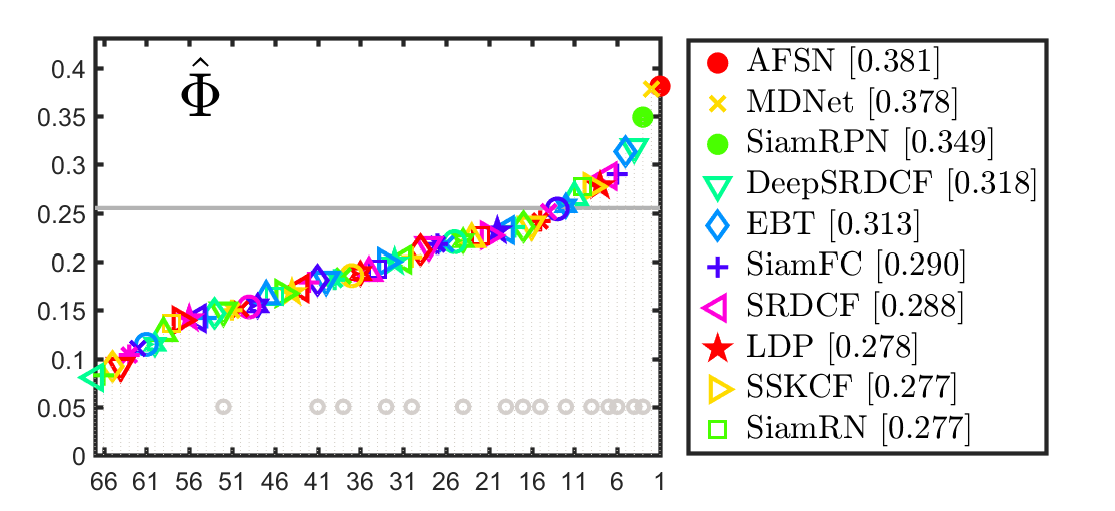}
\end{center}
  \caption{EAO scores rank on VOT2015}
\label{fig:VOT2015}
\end{figure}

 We compared our AFSN with 10 state-of-the-art trackers. The results are reported in Tab. \ref{tab:vot2015}. SiamFC and SiamRPN are added into comparisons as our baselines. As is shown in Fig. \ref{fig:VOT2015}, our tracker is able to rank the $1$st in EAO. SiamFC is one of the top trackers on VOT2015 which can run at frame rates beyond real time and achieves state-of-the-art performance. SiamRPN is able to gain 23\% increase in EAO, and our AFSN can achieve 0.381 in EAO, which is 9.2\% higher than SiamRPN. Also, AFSN is able to rank the $1$st in accuracy, the $2$nd in EFO and the $3$rd in failure. 

\begin{table}[htbp!]
    \centering
    \caption{Results about the state-of-the-art trackers in VOT2015. \textcolor{red}{Red}, \textcolor{blue}{blue} and \textcolor{green}{green} represent the $1$st, $2$nd and $3$rd respectively.}
    \begin{tabular}{ccccc}
    \hline
         \textbf{Tracker} & \textbf{EAO} & \textbf{Accuracy} & \textbf{Failure} & \textbf{EFO} \\
    \hline
    DeepSRDCF & \textcolor{green}{0.3181} & 0.56 & \textcolor{blue}{1.0} & 0.38 \\
    EBT & 0.313 & 0.45 & 1.02 & 1.76 \\
    SRDCF & 0.2877 & 0.55 & 1.18 & 1.99 \\
    LDP & 0.2785 & 0.49 & 1.3 & 4.36 \\
    sPST & 0.2767 & 0.54 & 1.42 & 1.01 \\
    SC-EBT & 0.2548 & 0.54 & 1.72 & 0.8 \\
    NSAMF & 0.2536 & 0.53 & 1.29 & 5.47 \\
    Struck & 0.2458 & 0.46 & 1.5 & 2.44 \\
    RAJSSC & 0.242 & \textcolor{green}{0.57} & 1.75 & 2.12 \\
    S3Tracker & 0.2403 & 0.52 & 1.67 & \textcolor{green}{14.27} \\ 
    SiamFC-3s & 0.2904 & 0.54 & 1.42 & 8.68 \\
    SiamRPN & \textcolor{blue}{0.349} & \textcolor{blue}{0.58} & \textcolor{red}{0.93} & \textcolor{red}{23.0} \\
    \hline
    \textbf{AFSN} & \textcolor{red}{0.381} & \textcolor{red}{0.59} & \textcolor{green}{1.01} & \textcolor{blue}{20.4} \\
    \hline
    \end{tabular}
    \label{tab:vot2015}
\end{table}

\subsection{Results on VOT2016}
\begin{figure}[htbp!]
\begin{center}
\includegraphics[width=.46\textwidth]{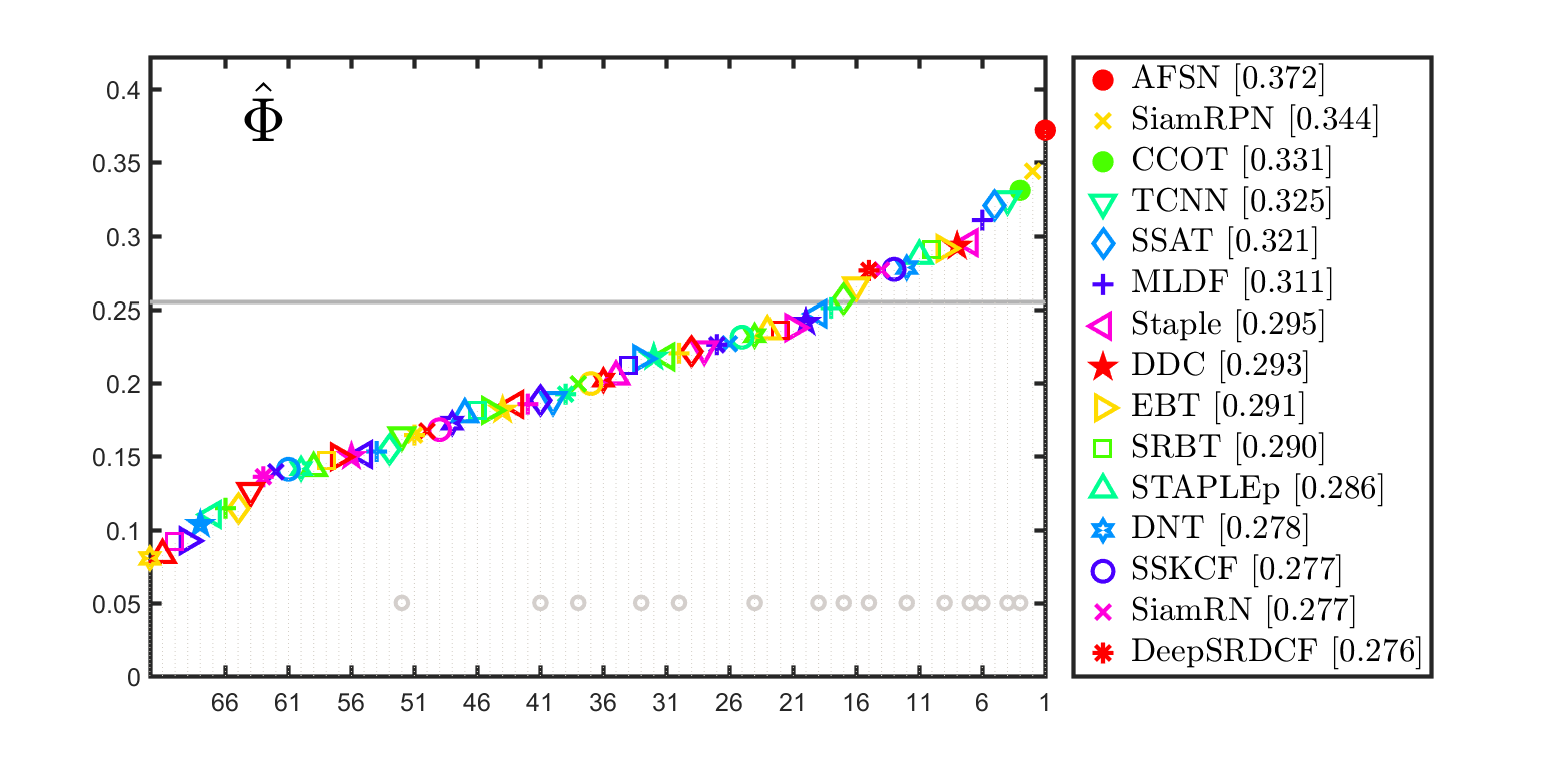}
\end{center}
  \caption{Expected overlap scores in the VOT2016 challenge.}
\label{fig:VOT2016result}
\end{figure}

The video sequences in VOT2016 are the same as VOT2015, but the groundtruth bounding boxes are re-annotated. We compare our trackers to the top 20 trackers in the challenge. As shown in Fig. \ref{fig:VOT2016result}, AFSN can outperform all entries in challenge. Tab. \ref{tab:vot2016} shows the information of several state-of-the-art trackers. AFSN can achieve a 12.4\% gain in EAO, 15.1\% in accuracy compared with C-COT \cite{ccot}. Our tracker also outperforms SiamRPN in EAO, accuracy and failure. Most prominently, Our tracker operates at 136 FPS, which is $425\times$  faster than C-COT.

\begin{table}[htbp!]
    \centering
    \caption{Results about the published state-of-the-art trackers in VOT2016. \textcolor{red}{Red}, \textcolor{blue}{blue} and \textcolor{green}{green} represent the $1$st, $2$nd and $3$rd respectively.}
    \begin{tabular}{ccccc}
    \hline
         \textbf{Tracker} & \textbf{EAO} & \textbf{Accuracy} & \textbf{Failure} & \textbf{EFO} \\
    \hline
    C-COT & \textcolor{green}{0.331} & 0.53 & \textcolor{red}{0.85} & 0.507 \\
    ECO-HC & 0.322 & 0.53 & 1.08 & \textcolor{green}{15.13} \\
    Staple & 0.2952 & 0.54 & 1.35 & 11.14 \\
    EBT & 0.2913 & 0.47 & \textcolor{blue}{0.9} & 3.011 \\
    MDNet & 0.257 & 0.54 & 1.2 & 0.534 \\
    SiamRN & 0.2766 & \textcolor{green}{0.55} & 1.37 & 5.44 \\
    SiamAN & 0.2352 & 0.53 & 1.65 & 9.21 \\
    SiamRPN & \textcolor{blue}{0.3441} & \textcolor{blue}{0.56} & 1.08 & \textcolor{red}{23.3} \\
    \hline
    AFSN & \textcolor{red}{0.372} & \textcolor{red}{0.61} & \textcolor{green}{1.04} & \textcolor{blue}{20.6} \\
    \hline
    \end{tabular}
    \label{tab:vot2016}
\end{table}

\subsection{Results on VOT2018}
VOT2018 \cite{vot2018} dataset consists of 60 video sequences. The performance is also evaluated in terms of accuracy (average overlap in the course of successful tracking) and robustness (failure rate). EAO is the combination of these two measurements. Tab. \ref{tab:Trackernet} shows the comparison of our approach with the top 10 trackers in the VOT2018 challenge. Among the top trackers, our AFSN achieves the best EAO and accuracy, while having competitive robustness. Although recently released SiamRPN++ \cite{SiamRPN++} can achieve 0.414 in EAO, our AFSN can operate at $3.9\times$  faster (136 FPS \textit{v.s.}35 FPS) than SiamRPN++ with only a 4\% drop in EAO.

\begin{figure}[htbp!]
\begin{center}
\includegraphics[width=.46\textwidth]{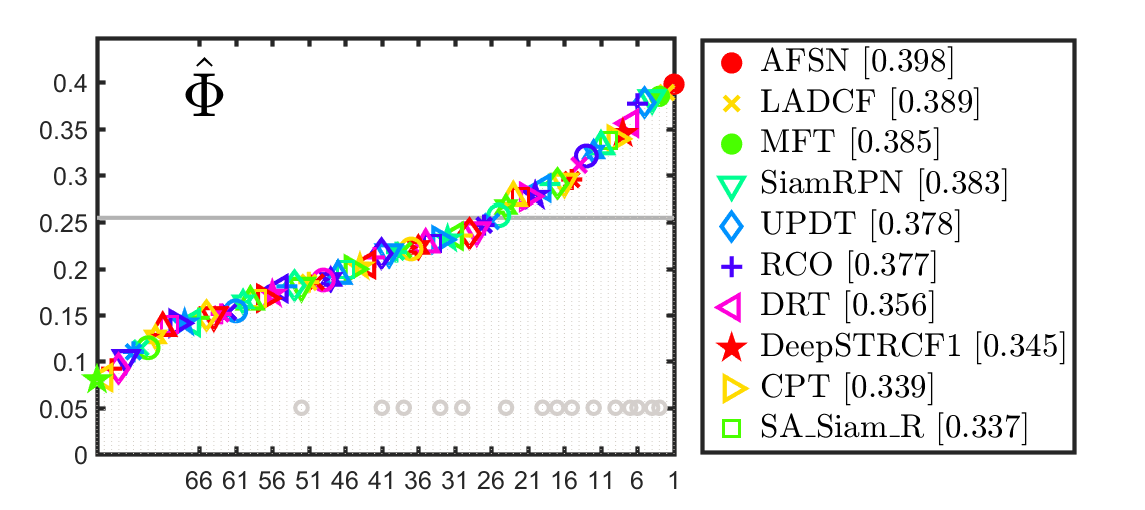}
\end{center}
  \caption{EAO scores rank on VOT2018}
\label{fig:VOT2018}
\end{figure}

\begin{table}[htbp!]
    \centering
    \caption{Results about the published state-of-the-art trackers in VOT2018. \textcolor{red}{Red}, \textcolor{blue}{blue} and \textcolor{green}{green} represent the $1$st, $2$nd and $3$rd respectively.}
    \begin{tabular}{cccc}
    \hline
         \textbf{Tracker} & \textbf{EAO} & \textbf{Accuracy} & \textbf{Robustness}\\
    \hline
    LADCF & \textcolor{blue}{0.389} & 0.503 & \textcolor{green}{0.159}\\
    MFT & \textcolor{green}{0.385} & 0.505 & \textcolor{red}{0.140}\\
    DaSiamRPN & 0.383 & \textcolor{blue}{0.586} & 0.276\\
    UPDT & 0.378 & 0.536 & 0.184\\
    RCO & 0.376 & 0.507 & \textcolor{blue}{0.155}\\
    DRT & 0.356 & 0.519 & 0.201\\
    DeepSTRCF & 0.345 & 0.523 & 0.215\\
    CPT & 0.339 & 0.506 & 0.239\\
    SASiamR & 0.337 & \textcolor{green}{0.566} & 0.258\\
    DLSTpp & 0.325 & 0.543 & 0.224\\
    \hline
    AFSN & \textcolor{red}{0.398} & \textcolor{red}{0.589} & 0.204\\
    \hline
    \end{tabular}
    \label{tab:vot2018}
\end{table}

\subsection{Results on TrackingNet}
TrackingNet \cite{TrackingNet} is the first large-scale dataset and benchmark for object tracking in the wild. It provides more than 30K videos with more than 14 million dense bounding box annotations sampled from YouTube. The dataset covers a wide selection of object classes in broad and diverse context. The trackers are evaluated using an online evaluation server on a test set of 511 videos. The results of precision, normalized precision and success are reported in Tab. \ref{tab:Trackernet}. MDNet achieves scores of 0.565 and UPDT achieves 0.611 in terms of precision and success, respectively. Our AFSN ranks the $1$st with relative gains of 7.4\% and 7.2\%. 

\begin{table}[htbp!]
    \centering
    \caption{Comparison pn the TrackingNet test set with the state-of-the-art trackers. \textcolor{red}{Red}, \textcolor{blue}{blue} and \textcolor{green}{green} represent the $1$st, $2$nd and $3$rd respectively.}
    \begin{tabular}{cccc}
    \hline
         \textbf{Tracker} & \textbf{Precision} & \textbf{Norm precision} & \textbf{Success}\\
    \hline
    UPDT & \textcolor{green}{0.557} & \textcolor{green}{0.702} & \textcolor{blue}{0.611}\\
    MDNet & \textcolor{blue}{0.565} & \textcolor{blue}{0.705} & \textcolor{green}{0.606}\\
    CFNet & 0.533 & 0.654 & 0.578\\
    SiamFC & 0.533 & 0.666 & 0.571\\
    DaSiamRPN & 0.413 & 0.602 & 0.568\\
    ECO & 0.492 & 0.618 & 0.554\\
    CSRDCF & 0.480 & 0.622 & 0.534\\
    SAMF & 0.477 & 0.598 & 0.504\\
    Staple & 0.470 & 0.603 & 0.528\\
    \hline
    AFSN & \textcolor{red}{0.607} & \textcolor{red}{0.738} & \textcolor{red}{0.655}\\
    \hline
    \end{tabular}
    \label{tab:Trackernet}
\end{table}
\section{Conclusions}
This paper presents the first in-depth study on anchor free object tracking called Anchor Free Siamese Network (AFSN). Unlike traditional Siamese trackers, the proposed AFSN does not need to enumerate an exhaustive list of potential object locations and classify each anchor. A target object is characterized by a bounding box center, tracking offset and object size. All three are regressed by Siamese network, performed one time per frame. We also optimize Siamese network architecture for AFSN, and perform extensive ablation experiments to quantitatively illustrate effectiveness of AFSN. We evaluate AFSN using five most commonly used benchmarks and compare to the best anchor-based trackers with source codes available for each benchmark. AFSN is $3\times-425\times$ faster than these best anchor based trackers. AFSN is also 5.97\% to 12.4\% more accurate in terms of all metrics for benchmark sets OTB2015, VOT2015, VOT2016, VOT2018 and TrackingNet, except that SiamRPN++ is 4\% better than AFSN in terms of Expected Average Overlap (EAO) on VOT2018 (but SiamRPN++ is $3.9\times$ slower).

{\small
\bibliographystyle{ieee_fullname}
\bibliography{main}

\begin{thebibliography}{10}\itemsep=-1pt

\bibitem{D2}
S. {Avidan}.
\newblock Support vector tracking.
\newblock {\em IEEE Transactions on Pattern Analysis and Machine Intelligence},
  26(8):1064--1072, Aug 2004.

\bibitem{D1}
S. {Avidan}.
\newblock Ensemble tracking.
\newblock {\em IEEE Transactions on Pattern Analysis and Machine Intelligence},
  29(2):261--271, Feb 2007.

\bibitem{staple}
L. Bertinetto, J. Valmadre, S. Golodetz, O. Miksik, and P.~H. Torr.
\newblock Staple: Complementary learners for real-time tracking.
\newblock In {\em The IEEE Conference on Computer Vision and Pattern
  Recognition (CVPR)}, June 2016.

\bibitem{SiamFC}
L. Bertinetto, J. Valmadre, J.~F. Henriques, A. Vedaldi, and P.~H. Torr.
\newblock Fully-convolutional siamese networks for object tracking.
\newblock In {\em Computer Vision -- ECCV 2016 Workshops}, pages 850--865,
  Cham, 2016. Springer International Publishing.

\bibitem{MOSSE}
D. Bolme, J. Beveridge, B. Draper, and Y. Lui.
\newblock Visual object tracking using adaptive correlation filters.
\newblock pages 2544--2550, 06 2010.

\bibitem{ACFN}
J. {Choi}, H.~J. {Chang}, S. {Yun}, T. {Fischer}, Y. {Demiris}, and J.~Y.
  {Choi}.
\newblock Attentional correlation filter network for adaptive visual tracking.
\newblock In {\em 2017 IEEE Conference on Computer Vision and Pattern
  Recognition (CVPR)}, pages 4828--4837, July 2017.

\bibitem{dsst}
M. Danelljan, G. Häger, and F. Khan.
\newblock Accurate scale estimation for robust visual tracking.
\newblock {\em British Machine Vision Conference}, pages 1--11, 01 2014.

\bibitem{ccot}
M. Danelljan, A. Robinson, F. Khan, and M. Felsberg.
\newblock Beyond correlation filters: Learning continuous convolution operators
  for visual tracking.
\newblock volume 9909, pages 472--488, 10 2016.

\bibitem{Doellinger}
J. {Doellinger}, V.~S. {Prabhakaran}, L. {Fu}, and M. {Spies}.
\newblock Environment-aware multi-target tracking of pedestrians.
\newblock {\em IEEE Robotics and Automation Letters}, 4(2):1831--1837, April
  2019.

\bibitem{D3}
H. Grabner, M. Grabner, and H. Bischof.
\newblock Real-time tracking via on-line boosting.
\newblock volume~1, pages 47--56, 01 2006.

\bibitem{DSiam}
Q. {Guo}, W. {Feng}, C. {Zhou}, R. {Huang}, L. {Wan}, and S. {Wang}.
\newblock Learning dynamic siamese network for visual object tracking.
\newblock In {\em 2017 IEEE International Conference on Computer Vision
  (ICCV)}, pages 1781--1789, Oct 2017.

\bibitem{Struck}
S. {Hare}, A. {Saffari}, and P.~H.~S. {Torr}.
\newblock Struck: Structured output tracking with kernels.
\newblock In {\em 2011 International Conference on Computer Vision}, pages
  263--270, Nov 2011.

\bibitem{DLtrack1}
D. Held, S. Thrun, and S. Savarese.
\newblock Learning to track at 100 fps with deep regression networks.
\newblock In Bastian Leibe, Jiri Matas, Nicu Sebe, and Max Welling, editors,
  {\em Computer Vision -- ECCV 2016}, pages 749--765, Cham, 2016. Springer
  International Publishing.

\bibitem{CSK}
J. Henriques, R. Caseiro, P. Martins, and J. Batista.
\newblock Exploiting the circulant structure of tracking-by-detection with
  kernels.
\newblock volume 7575, pages 702--715, 10 2012.

\bibitem{KCF}
J.~F. {Henriques}, R. {Caseiro}, P. {Martins}, and J. {Batista}.
\newblock High-speed tracking with kernelized correlation filters.
\newblock {\em IEEE Transactions on Pattern Analysis and Machine Intelligence},
  37(3):583--596, March 2015.

\bibitem{CNNSVM}
S. Hong, T. You, S. Kwak, and B. Han.
\newblock Online tracking by learning discriminative saliency map with
  convolutional neural network.
\newblock In {\em Proceedings of the 32Nd International Conference on
  International Conference on Machine Learning - Volume 37}, ICML'15, pages
  597--606. JMLR.org, 2015.

\bibitem{GOT10k}
L.H. Huang, X. Zhao, and K.Q. Huang.
\newblock Got-10k: A large high-diversity benchmark for generic object tracking
  in the wild.
\newblock {\em ArXiv}, abs/1810.11981, 2018.

\bibitem{G1}
X. {Jia}, H. {Lu}, and M. {Yang}.
\newblock Visual tracking via adaptive structural local sparse appearance
  model.
\newblock In {\em 2012 IEEE Conference on Computer Vision and Pattern
  Recognition}, pages 1822--1829, June 2012.

\bibitem{vot2016}
M. Kristan, A. Leonardis, J. Matas, M. Felsberg, R. Pflugfelder, and L.
  {\v{C}}ehovin.
\newblock The visual object tracking vot2016 challenge results.
\newblock In {\em Computer Vision -- ECCV 2016 Workshops}, pages 777--823,
  Cham, 2016. Springer International Publishing.

\bibitem{vot2018}
M. Kristan, A. Leonardis, J. Matas, M. Felsberg, R. Pflugfelder, L. Zajc, T.
  Vojir, G. Bhat, and A. Luke{\v{z}}i{\v{c}}.
\newblock The sixth visual object tracking vot2018 challenge results.
\newblock In {\em Computer Vision -- ECCV 2018 Workshops}, pages 3--53, Cham,
  2019. Springer International Publishing.

\bibitem{vot2017}
M. {Kristan}, A. {Leonardis}, J. {Matas}, M. {Felsberg}, R. {Pflugfelder},
  L.~C. {Zajc}, T. {Vojír}, G. {Häger}, A. {Lukežic}, A. {Eldesokey}, G.
  {Fernández}, and Á. {García-Martín}.
\newblock The visual object tracking vot2017 challenge results.
\newblock In {\em 2017 IEEE International Conference on Computer Vision
  Workshops (ICCVW)}, pages 1949--1972, Oct 2017.

\bibitem{vot2015}
M. Kristan, J. Matas, A. Leonardis, M. Felsberg, L. Cehovin, G. Fernandez, T.
  Vojir, G. Hager, G. Nebehay, and R. Pflugfelder.
\newblock The visual object tracking vot2015 challenge results.
\newblock In {\em The IEEE International Conference on Computer Vision (ICCV)
  Workshops}, December 2015.

\bibitem{CornerNet}
H. Law and J. Deng.
\newblock Cornernet: Detecting objects as paired keypoints.
\newblock {\em International Journal of Computer Vision}, Aug 2019.

\bibitem{SiamRPN++}
B. Li, W. Wu, Q. Wang, F.Y. Zhang, J.L. Xing, and J.J. Yan.
\newblock Siamrpn++: Evolution of siamese visual tracking with very deep
  networks.
\newblock {\em CoRR}, abs/1812.11703, 2018.

\bibitem{SiamRPN}
B. Li, J.J. Yan, W. Wu, Z. Zhu, and X.L. Hu.
\newblock High performance visual tracking with siamese region proposal
  network.
\newblock In {\em The IEEE Conference on Computer Vision and Pattern
  Recognition (CVPR)}, June 2018.

\bibitem{focal}
T.Y. Lin, P. Goyal, R. Girshick, K.M. He, and P. Dollar.
\newblock Focal loss for dense object detection.
\newblock In {\em The IEEE International Conference on Computer Vision (ICCV)},
  Oct 2017.

\bibitem{Liu2019}
X.C. Liu, H.D. Ma, and S.Q. Li.
\newblock Pvss: A progressive vehicle search system for video surveillance
  networks.
\newblock {\em Journal of Computer Science and Technology}, 34(3):634--644, May
  2019.

\bibitem{G2}
X. Mei and H.B. Ling.
\newblock Robust visual tracking using l(1) minimization.
\newblock pages 1436 -- 1443, 11 2009.

\bibitem{G3}
X. Mei, H.B. Ling, Y. Wu, E. Blasch, and l. Bai.
\newblock Minimum error bounded efficient l1 tracker with occlusion detection
  (preprint).
\newblock pages 1257--1264, 06 2011.

\bibitem{TrackingNet}
M. Muller, A. Bibi, S. Giancola, S. Alsubaihi, and B. Ghanem.
\newblock Trackingnet: A large-scale dataset and benchmark for object tracking
  in the wild.
\newblock In {\em The European Conference on Computer Vision (ECCV)}, September
  2018.

\bibitem{MDNet}
H. {Nam} and B. {Han}.
\newblock Learning multi-domain convolutional neural networks for visual
  tracking.
\newblock In {\em 2016 IEEE Conference on Computer Vision and Pattern
  Recognition (CVPR)}, pages 4293--4302, June 2016.

\bibitem{ExtremeNet}
F. Nasse and G.~A. Fink.
\newblock A bottom-up approach for learning visual object detection models from
  unreliable sources.
\newblock In Axel Pinz, Thomas Pock, Horst Bischof, and Franz Leberl, editors,
  {\em Pattern Recognition}, pages 488--497, Berlin, Heidelberg, 2012. Springer
  Berlin Heidelberg.

\bibitem{ThreeSigma}
F. Pukelsheim.
\newblock The three sigma rule.
\newblock {\em The American Statistician}, 48(2):88--91, 1994.

\bibitem{Fasterrcnn}
S.Q. Ren, K.M. He, R. Girshick, and J. Sun.
\newblock Faster r-cnn: Towards real-time object detection with region proposal
  networks.
\newblock In C. Cortes, N.~D. Lawrence, D.~D. Lee, M. Sugiyama, and R. Garnett,
  editors, {\em Advances in Neural Information Processing Systems 28}, pages
  91--99. Curran Associates, Inc., 2015.

\bibitem{ALOV}
A.~W.~M. {Smeulders}, D.~M. {Chu}, R. {Cucchiara}, S. {Calderara}, A.
  {Dehghan}, and M. {Shah}.
\newblock Visual tracking: An experimental survey.
\newblock {\em IEEE Transactions on Pattern Analysis and Machine Intelligence},
  36(7):1442--1468, July 2014.

\bibitem{cf2}
C. Sun, D. Wang, H.C. Lu, and M.H. Yang.
\newblock Correlation tracking via joint discrimination and reliability
  learning.
\newblock 04 2018.

\bibitem{SINT}
R. Tao, E. Gavves, and A. Smeulders.
\newblock Siamese instance search for tracking.
\newblock pages 1420--1429, 06 2016.

\bibitem{G4}
D. {Wang}, H. {Lu}, and M. {Yang}.
\newblock Online object tracking with sparse prototypes.
\newblock {\em IEEE Transactions on Image Processing}, 22(1):314--325, Jan
  2013.

\bibitem{Attentional}
Q. {Wang}, Z. {Teng}, J. {Xing}, J. {Gao}, W. {Hu}, and S. {Maybank}.
\newblock Learning attentions: Residual attentional siamese network for high
  performance online visual tracking.
\newblock In {\em 2018 IEEE/CVF Conference on Computer Vision and Pattern
  Recognition}, pages 4854--4863, June 2018.

\bibitem{DONOTLOSE}
Q. Wang, M.D. Zhang, J.L. Xing, J. Gao, W.M. Hu, and Steve M.
\newblock Do not lose the details: Reinforced representation learning for high
  performance visual tracking.
\newblock In {\em Proceedings of the Twenty-Seventh International Joint
  Conference on Artificial Intelligence, {IJCAI-18}}, pages 985--991.
  International Joint Conferences on Artificial Intelligence Organization, 7
  2018.

\bibitem{otb2013}
Y. Wu, J. Lim, and M.H. Yang.
\newblock Online object tracking: A benchmark.
\newblock In {\em The IEEE Conference on Computer Vision and Pattern
  Recognition (CVPR)}, June 2013.

\bibitem{otb15}
Y. {Wu}, J. {Lim}, and M. {Yang}.
\newblock Object tracking benchmark.
\newblock {\em IEEE Transactions on Pattern Analysis and Machine Intelligence},
  37(9):1834--1848, Sep. 2015.

\bibitem{SiamFT}
X.~C. Zhang, P. Ye, S.~Y. Peng, J. Liu, K. Gong, and G. Xiao.
\newblock Siamft: An rgb-infrared fusion tracking method via fully
  convolutional siamese networks.
\newblock {\em IEEE Access}, 7:122122--122133, 2019.

\bibitem{DSiamMFT}
X.~C. Zhang, P. Ye, S.~Y. Peng, J. Liu, and G. Xiao.
\newblock Dsiammft: An rgb-t fusion tracking method via dynamic siamese
  networks using multi-layer feature fusion.
\newblock {\em Signal Processing: Image Communication}, 84:115756, 2020.

\bibitem{SiamDW}
Z.P. Zhang, H.W. Peng, and Q. Wang.
\newblock Deeper and wider siamese networks for real-time visual tracking.
\newblock {\em CoRR}, abs/1901.01660, 2019.

\bibitem{G5}
W. {Zhong}, H. {Lu}, and M. {Yang}.
\newblock Robust object tracking via sparse collaborative appearance model.
\newblock {\em IEEE Transactions on Image Processing}, 23(5):2356--2368, May
  2014.

\bibitem{CenterNet}
X.Y. Zhou, D.Q. Wang, and P. Kr{\"{a}}henb{\"{u}}hl.
\newblock Objects as points.
\newblock {\em CoRR}, abs/1904.07850, 2019.

\bibitem{DaSiamRPN}
Z. Zhu, Q. Wang, B. Li, W. Wu, J.J. Yan, and W.M. Hu.
\newblock Distractor-aware siamese networks for visual object tracking.
\newblock In {\em The European Conference on Computer Vision (ECCV)}, September
  2018.

\end{thebibliography}
}

\end{document}